\documentclass[12pt]{article}
\usepackage[utf8]{inputenc}
\usepackage{a4wide}
\usepackage{BunchOfCommands}
\usepackage{subfiles}

\title{Neural Langevin Dynamics: towards interpretable Neural Stochastic Differential Equations}
\author{Simon Koop, Mark Peletier, Jim Portegies, Vlado Menkovski}

\begin{document}
\bibliographystyle{plainnat}
\maketitle
\begin{abstract}
    Neural Stochastic Differential Equations (NSDE) have been trained as both Variational Autoencoders, and as GANs. However, the resulting Stochastic Differential Equations can be hard to interpret or analyse due to the generic nature of the drift and diffusion fields. By restricting our NSDE to be of the form of Langevin dynamics, and training it as a VAE, we obtain NSDEs that lend themselves to more elaborate analysis and to a wider range of visualisation techniques than a generic NSDE. More specifically, we obtain an energy landscape, the minima of which are in one-to-one correspondence with latent states underlying the used data. This not only allows us to detect states underlying the data dynamics in an unsupervised manner, but also to infer the distribution of time spent in each state according to the learned SDE. More in general, restricting an NSDE to Langevin dynamics enables the use of a large set of tools from computational molecular dynamics for the analysis of the obtained results.
\end{abstract}

\section{Introduction}
In recent years there has been widespread interest in Neural SDEs for generative modelling. Neural SDEs (NSDEs) learn the parameters in stochastic differential equations, which are natural extensions of ordinary differential equations to situations where uncertainty arises from many small and unobserved interactions, such as stock prices and molecular dynamics \cite{li2020scalable, kidger2021sdegan}. As such, Neural SDEs can be seen as incorporating uncertainty in Neural Ordinary Differential Equations. 

NSDEs are said to combine the strengths of deep learning and classical modelling \cite{kidger2021sdegan, hasan2022identifying}. Deep learning brings performance and capacity, whereas classical modelling derives a lot of value from its interpretability. However, although from some perspective the the dynamical system learned by an NSDE is itself a classical model, it is still a very complicated object, which makes it questionable whether NSDEs actually provide interpretability. 

Indeed, what is often overlooked in Neural Differential Equation literature, is that finding the differential equation is only the start of the process. For interpretation, one needs to analyse stability, find stationary solutions, and analyse asymptotic behaviour, to name a few. In the particular case of NSDEs, one would want to find basins of attraction, as they could correspond to discrete states in the dynamics, would like to extract how long the system spends in each state and at what rates transitions occur. For general dynamical systems, these tasks are prohibitively complicated, which severely hampers interpretability.

To make NSDEs more interpretable, we introduce Neural Langevin Dynamics (NLD), in which we replace the generic vector field, which serves as the drift, by a gradient field, and let the system evolve by Langevin dynamics. This has the following advantages.

For one, there is a direct interpretation of ``states'' in the original system as the basins of attraction around local minima of the energy (e.g.~\cite{lelievre2010free}) These local minima can be found using the vast toolbox of scalar optimisation methods --- e.g. \cite{nocedal2006numerical, lee2016gradient, zhang2021escape, jin2017escape}. 

\begin{figure}
    \centering
    \includemaybesvg[.45\textwidth]{../images/contour.pdf}{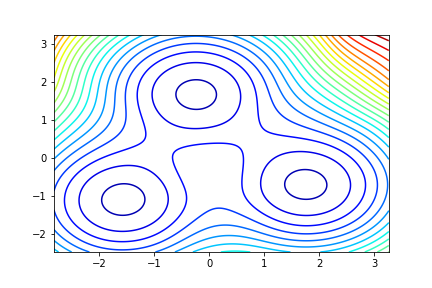}
    \caption{Contour of the energy landscape of an NLD model.\label{fig:contour:intro}}
\end{figure}

Sub-level sets of the energy function can then be used as approximations of the basins of attraction. This is illustrated by e.g.~\Cref{fig:contour:intro}, in which the contour lines clearly define regions that belong to specific states.

Moreover, transitions between local minima of the energy can be characterised in terms of mountain-pass connections between them. Such connections can be found using the numerical Mountain-Pass method~\cite{ChoiMcKenna93}, also called String Method~\cite{WeinanRenVanden-Eijnden02,WeinanERenVanden-Eijnden05}, and the corresponding transition rates can be estimated using Kramers' formula~\cite{Kramers40,BovierEckhoffGayrardKlein04,LandimMarianiSeo19}.

Finally, there exist many more tools for visualising scalar functions than for visualising vector fields, aiding in the understanding by users of these models. 

In this work, after introducing NLD in more detail, we show its potential and validity by training it as a Variational Autoencoder (VAE) on a pair of example problems. We show that the local minima indeed coincide with the true states underlying the data, and that information on relative occurrence of states can be extracted from the learned energy landscape.

Our results indicate that NLD allows us to extract interpretable information about the data dynamics, and in such a way that it opens up the use of tools from optimization and analysis, effectively providing a bridge between deep learning and classical modelling. 

\section{Related Work}
\cite{tzen2019neural, peluchetti2019infinitely} view NSDEs as infinitely deep neural networks but don't use them as time series models. \cite{kidger2021sdegan} train their NSDE as a GAN instead of as a VAE. \cite{li2020scalable, kidger2021efficient} extend the so called adjoint sensitivity method to NSDEs and provide algorithms for storing Wiener processes. We do not use the adjoint sensitivity method but do use their algorithms for training NSDEs in the form of the torchsde package ---\url{https://github.com/google-research/torchsde}.  Moreover, \cite{li2020scalable} introduce the training of an NSDE as a VAE on sequential data making them the most closely related to this work. \cite{hasan2022identifying} also train NSDEs as VAEs but focus mostly on the dimensionality of the data space.

NSDEs can be viewed in the wider context of neural differential equations. The seminal paper in this area was \cite{chen2018neural}. These were extended to models for sequential data in, among others \cite{rubanova2019latent,kidger2020neural, yildiz2019ode2vae}. 

Moreover, NLD can be viewed in a line of physics inspired neural network models such as \cite{cranmer2020lagrangian, greydanus2019hamiltonian, toth2020hamiltonian, botev2021priors}. The focus in the cited works however was more on performance and theoretical generalisability whereas our work focuses more on interpretability. 

Langevin dynamics themselves have been extensively researched since the start of the 20th century and remain a topic of active research \cite{Kramers40, berglund2011kramers, markowich2000trend}. Besides their relevance for physical modelling, they play an important role in MCMC sampling see e.g. \cite{lelievre2010free}.

\section{Method}
\label{sec:method}
\subsection{Neural Stochastic Differential Equations}
In general, when training Neural Stochastic Differential Equations as Variational Auto-Encoders on some data $X=\left\{x =\{x_t\}_{=0}^T \right\}$, one has two stochastic differential equations parameterising the prior $p_\theta(z)$ and approximate posterior $q_\phi(z\mid x)$ distributions:
\begin{align}
    p_\theta(z) \sim \dif z_t &= h_\theta(z_t, t)\dif t + g_\theta(z_t, t)\dif W_t\\&\text{(prior)},\label{eq:general:p}\\
    q_\phi(z\mid x) \sim \dif z_t &= f_\phi(z_t, t, c_t)\dif t + g_\theta(z_t, t)\dif W_t\\&\text{(approximate posterior)}\label{eq:general:q}.
\end{align}
Here $c_t$ is a function of the data $x$ obtained by some encoder architecture, in this work a GRU. Throughout this work, $c_t$ will only depend on ${x_s}_{s\leq t}$ so the models all operate in an online fashion. The network being integrated against $t$, i.e. $h_\theta$ in the prior and $f_\phi$ in the approximate posterior, is called the drift, and the network being integrated against the Wiener Process, i.e. $g_\theta$, is called the diffusion.

As shown in \cite{li2020scalable}, the Kullback-Leibler divergence between the approximate posterior and the prior is then given by 
\begin{align}
    \dkl\left(q_\phi(z\mid x)\mid\mid p(z)\right) &= \half\int_0^T |u(z_t, t, c_t)|^2\dif t,\\
    g_\theta(z_t, t)\, u(z_t, t, c_t) &= f_\phi(z_t, t, c_t) - h_\theta(z_t, t),
\end{align}
where $T$ is the final time. 

Note that $p_\theta(z)$ is a learned distribution, parameterised by a neural network, representing the latent dynamics of the dataset as a whole. However, it is not easy to extract useful information about these latent dynamics from this representation of the distribution due to the generality of \Cref{eq:general:p}. 

\subsection{Neural Langevin Dynamics}
We introduce Neural Langevin Dynamics as a more interpretable alternative to NSDE. In NLD, we replace the drift $h_\theta$ in \cref{eq:general:p} by the scaled gradient of an energy $E_\theta : \mathbb{R}^d \to \mathbb{R}$ that is given by a neural network, and modify the evolution to either overdamped or underdamped Langevin dynamics.

In the case of overdamped Langevin dynamics, the prior and posterior evolutions are determined by
\begin{align}
\dif z_t ={}& -\gamma\inv \nabla_z E_\theta(z_t) \dif t + \sqrt{2\beta\inv\gamma\inv}\dif W_t\\
\dif z_t ={}& - \gamma\inv \nabla_z E_\theta(z_t)\dif t \\&{}\sns + f_\phi(z_t, t, c_t)\dif t + \sqrt{2\beta\inv\gamma\inv}\dif W_t
\end{align}
whereas in the case of underdamped Langevin dynamics, they are determined by
\begin{align}
&\begin{cases} \dif z^q_t ={}& M\inv z^p_t\dif t\\\dif z^p_t ={}& \left[-\nabla E_\theta(z^q_t) -\gamma M\inv z^p_t\right]\dif t \\&+ \sqrt{2\gamma\beta\inv }\dif W_t \end{cases}\\
&\begin{cases} \dif z^q_t =&\sns M\inv z^p_t\dif t\\\dif z^p_t =&\sns - \left[\nabla E_\theta(z^q_t) + \gamma M\inv z^p_t\right]\dif t \\&\sns+ f_\phi(z^q_t, z^p_t, t, c_t) \dif t + \sqrt{2\gamma\beta\inv }\dif W_t. \end{cases}
\end{align} 
Here, $M$ is a positive diagonal matrix representing mass. The constants $M, \gamma$ and $\beta$ can either fixed constants, or can be learned while training the model.

In the underdamped case, $z^q$ plays the role of location, and $z^p$ plays the role of momentum. Only the location, $z^q$ should be used for decoding. Note that $z^q$ follows a second-order stochastic differential equation. 

One of the great advantages of using Langevin dynamics is that this way, the SDE of the prior distribution has a known stationary distribution of the form
\begin{align}
    p_\infty(z) &\sim \frac{\exp(-\beta E_\theta(z))}{\int_{\rel^d} \exp(-\beta E_\theta(y))\dif y}\\
    \intertext{in the overdamped case, and}
    p_\infty(z_q, z_p) &\sim \frac{\exp(-\beta E_\theta(z_q))}{\int_{\rel^d} \exp(-\beta E_\theta(y))\dif y} \otimes \mathcal{N}(z_p; 0, M\inv)
\end{align}
in the underdamped case. Due to ergodicity, the proportion of time spent in a region of the latent space, in the long run, corresponds to the mass attributed to that region by the stationary distribution, which only relies on the $\beta$ and $E_\theta$.

\begin{figure}
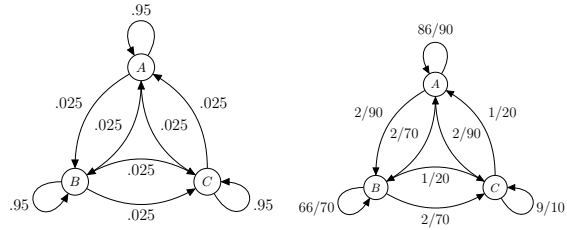

\centering
\begin{subfigure}[t]{.23\textwidth}
    \centering
    \includegraphics[width=\textwidth]{../tikz/symmetric_graph.tex}
    \caption{Graph for the first experiment.\label{fig:graph-1}}
\end{subfigure}
\begin{subfigure}[t]{.23\textwidth}
    \centering
    \includegraphics[width=\textwidth]{../tikz/asymmetric_graph.tex}
    \caption{Graph for the second experiment.\label{fig:graph-2}}
\end{subfigure}
\caption{Graphs of the Markov processes underlying Experiment 1 and 2.\label{fig:graphs}}
\end{figure}

\section{Experimental setup}
We evaluate the capability of our method to develop an interpretable representation of  data dynamics by training a model on data coming from a Markov chain. We then analyse the energy landscape of this model to recover the discrete states of the Markov chain and infer the time spent in each state. 

The data is generated as random-walks on a graph with three nodes. At each time step, a signal is emitted as a random vector sampled from a 15-dimensional multivariate normal distribution with mean and covariance matrix specified by the node the random walk is at. 

For the first experiment, the transition probability from any state to any different state is $0.025$ at each time step, resulting in a stationary distribution $(1/3,\, 1/3,\, 1/3)$, and an expected transition time of $20$ time-steps. An overview of this is shown in \Cref{fig:graph-1}.

For the second experiment, the transition probabilities are shown in \Cref{fig:graph-2}. In this experiment, the stationary distribution is $(.45,\, .35,\, .2)$. This means that long time after initialisation, the system is expected to spend $45\%$ of the time in state $A$, $35\%$ in state $B$, and $20\%$ of the time in state $C$. 

Code for both the generation of the datasets and the training of the models is provided in \url{https://anonymous.4open.science/r/dynvae-7F0D/}.

\begin{figure}
    \centering
    \begin{subfigure}[t]{.23\textwidth}
    \centering
    \includemaybesvg[\textwidth]{../images/quiver_2d.pdf}{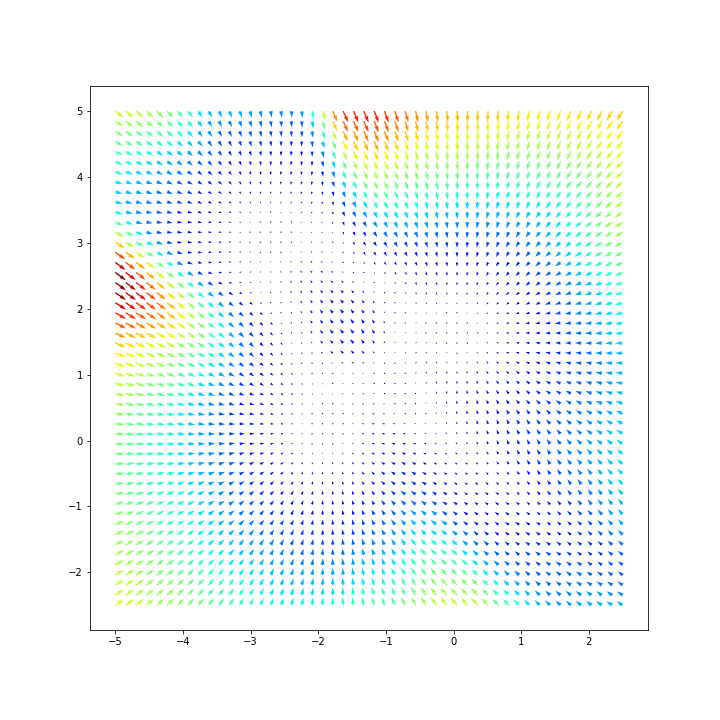}
    \caption{Quiver plot drift field.\label{fig:ngr-0:quiver}}
    \end{subfigure}
    \begin{subfigure}[t]{.23\textwidth}
    \centering
    \includemaybesvg[\textwidth]{../images/flow_lines_2d.pdf}{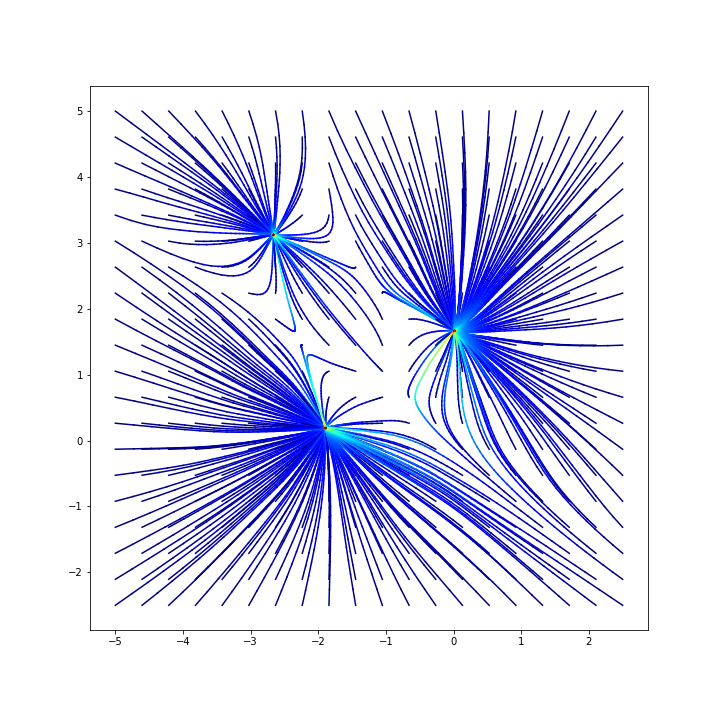}
    \caption{Flow lines drift field.\label{fig:ngr-0:flow-lines}}
    \end{subfigure}
    \begin{subfigure}[t]{.23\textwidth}
    \centering
    \includemaybesvg[\textwidth]{../images/quiver_3d.pdf}{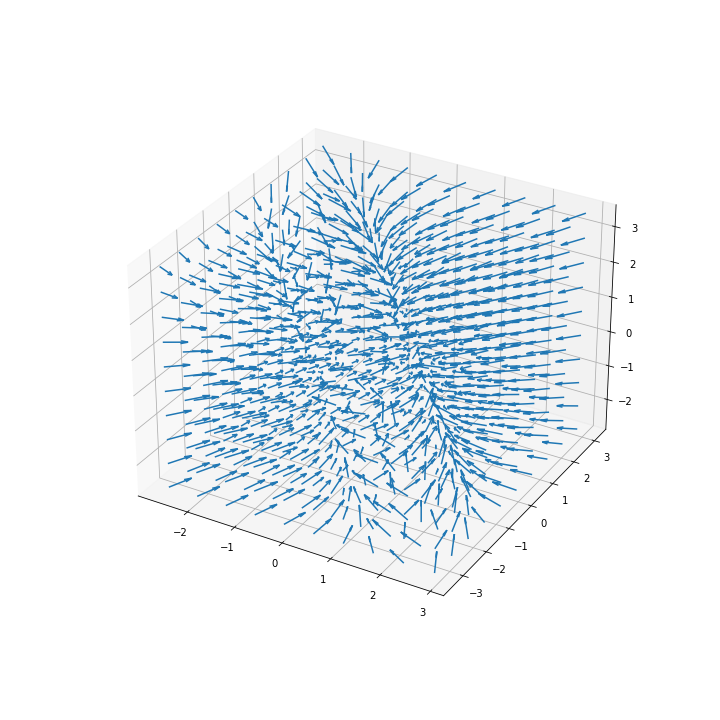}
    \caption{Quiverplot of 3d drift field. \label{fig:3d:quiver}}
    \end{subfigure}
    \begin{subfigure}[t]{.23\textwidth}
    \centering
    \includemaybesvg[\textwidth]{../images/flow_lines_3d_to_2d.pdf}{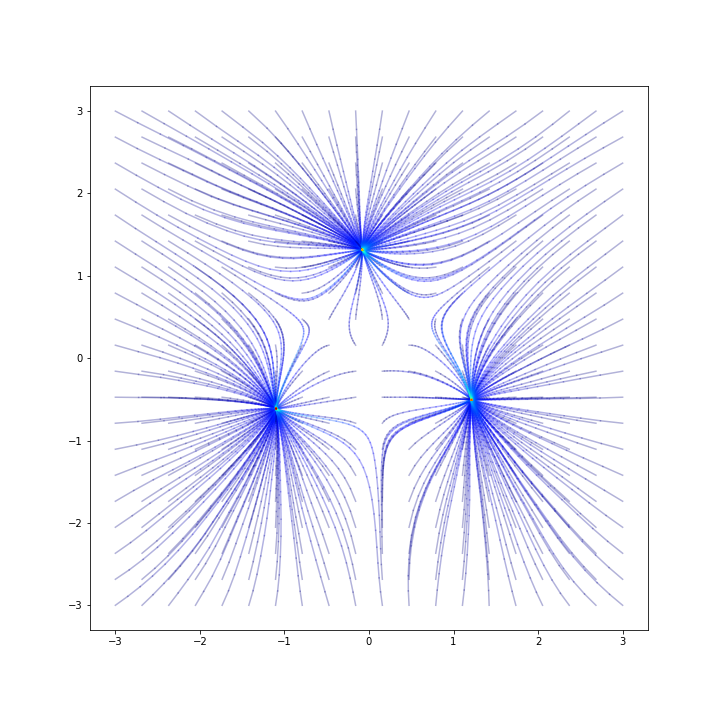}
    \caption{Projection of flowlines starting in a 2d plane onto that same plane. \label{fig:3d:flow}}
    \end{subfigure}    
    \begin{subfigure}[t]{.23\textwidth}
    \centering
    \includemaybesvg[\textwidth]{../images/energy_landscape_3d_plot.pdf}{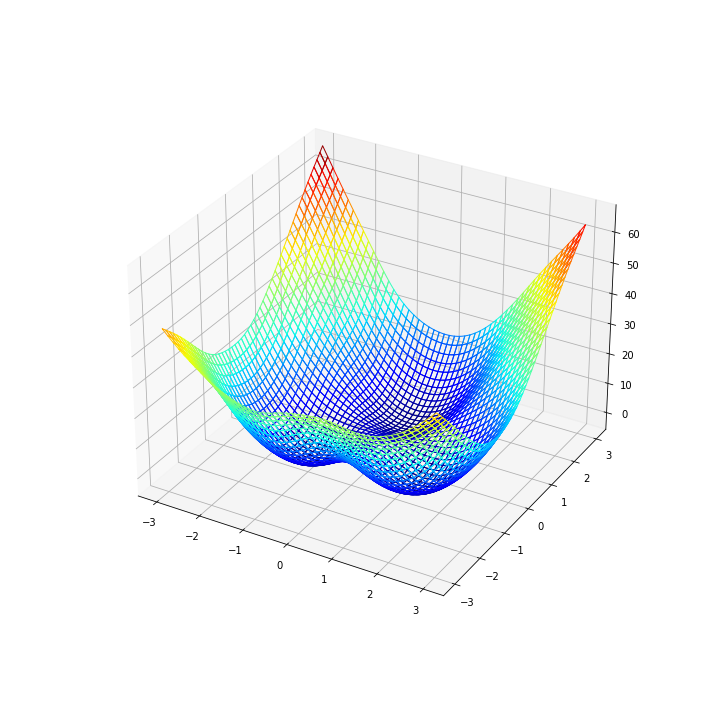}
    \caption{A 3d plot of the energy landscape restricted to a plane through the three minima. \label{fig:3d:3d}}
    \end{subfigure}
    \begin{subfigure}[t]{.23\textwidth}
    \centering
    \includemaybesvg[\textwidth]{../images/energy_landscape_contour_plot_2d.pdf}{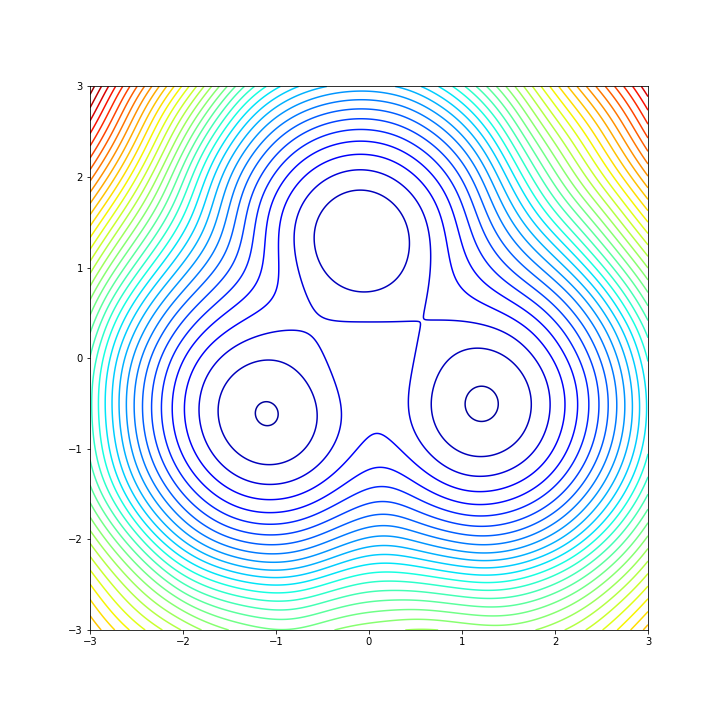}
    \caption{A contour plot of the energy restricted to the plane through the three minima. \label{fig:3d:contour:2d}}
    \end{subfigure}
    \centering
    \begin{subfigure}[t]{.23\textwidth}
    \centering
    \includemaybesvg[\textwidth]{../images/energy_landscape_contour_plot_3d.pdf}{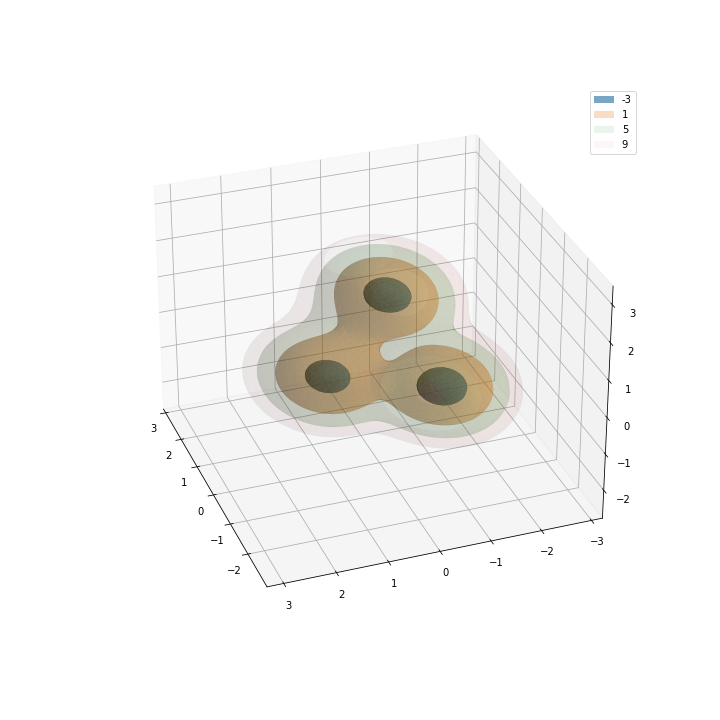}
    \caption{A 3d contour plot of the energy landscape.\label{fig:3d:contour:3d}}
    \end{subfigure}
    \caption{Visualisations of an NSDE (a-b) respectively an NLD (c-g) trained as a VAE on the data from experiment 1.}
    \label{fig:nsde-viz}
\end{figure}

\section{Visualisation advantages of NLD}
Visualisation is an important tool for obtaining human knowledge in general, and can be useful for extracting knowledge from a trained model.
To visualise trained models, if we only have a drift field, as is the case for a general NSDE, all we can really do is make a quiver plot of the drift field or plot the flow lines of the drift. For the two-dimensional case, this is shown in \Cref{fig:ngr-0:quiver} and \Cref{fig:ngr-0:flow-lines}. 

As shown in  \Cref{fig:3d:quiver}, a quiver plot of a three-dimensional vector field is already rather hard to interpret. One can restrict these visualisation techniques to a plane to make them more understandable, as is shown for a flow-line plot in \Cref{fig:3d:flow}. 

If on the other hand the drift field of the model is a gradient field of an energy function, as is the case with the NLD models, we can visualise the the energy function directly. If the dimensionality of the vectorfield is too high to do this directly, we can restrict ourselves to a two-dimensional plane within the larger space, as is shown in \Cref{fig:3d:3d} and \Cref{fig:3d:contour:2d}, or to a three-dimensional subspace, resulting in a visualisation like the one in \Cref{fig:3d:contour:3d}.

These additional tools can aid a practitioner in interpreting what the model has learned and translating the obtained results to either existing or new knowledge about the data.

\begin{table*}[!h]
    \centering
    \begin{tabular}{l|c|c|c}
         & From sampling (method \ref{e1a:sample})& $0^{\text{th}}$ order (method \ref{e1a:zero}) & $2^{\text{nd}}$ order (method \ref{e1a:2nd})\\
         \hline
         Overdamped & $0.047\pm 0.037$ & $0.051 \pm 0.022$ & $0.040\pm 0.013$\\
         Underdamped & $0.072\pm 0.068 $&$ 0.041 \pm0.029$ & $0.021\pm0.0044$
    \end{tabular}
    \caption{$\ell^1$ distance between estimated distribution of states and actual distribution of states in experiment 1.}
    \label{tab:states:1}
\end{table*}

\begin{table*}[!h]
    \centering
    \begin{tabular}{l|c|c|c}
        & From sampling & $0^{\text{th}}$ order & $2^{\text{nd}}$ order\\
        \hline
        Overdamped & $0.329 \pm 0.013$& $0.322 \pm 0.022$ & $0.324\pm0.017$  \\
        Underdamped & $0.335 \pm 0.027$ & $0.333\pm0.011$ & $0.336\pm 0.006$ 
    \end{tabular}
    \caption{First coordinate of the various estimations of the distribution of the state in experiment 1.\label{tab:coord:1}}
    
\end{table*}

\begin{table*}[!h]
    \centering
    \begin{tabular}{l|c|c}
         & mean accuracy & standard deviation\\
         \hline
         overdamped & $92.6\% \pm 0.6\%$ & $1.6\% \pm 0.2\%$\\
         underdamped & $82.2\% \pm 5.6\%$ & $3.6\% \pm 1.4\%$
    \end{tabular}
    \caption{Accuracy of unsupervised sequence segmentation for the same models as in \Cref{tab:states:1}\label{tab:seq-seg:1}}
\end{table*}

\section{Results}
We evaluate whether the learned energy landscape accurately represents the distribution of states underlying the data. 

\begin{figure}
    \centering
    \includemaybesvg[.5\textwidth]{../images/seqseg_online.pdf}{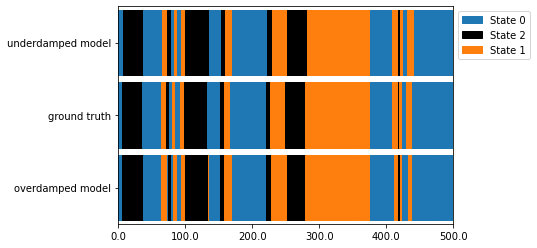}
    \caption{Results of unsupervised sequence segmentation of the first test sequence by both an underdamped and an overdamped NLD model.\label{fig:seq-seg:1}}
\end{figure}
\subsection{Experiment 1}
In the first experiment the data-generating Markov chain has an invariant distribution of $(1/3,1/3,1/3)$, and we investigate whether this distribution can be recovered from the prior in the trained Neural Langevin Dynamics network. 

We estimate the distribution of states in the learned prior in three different ways:
\begin{enumerate}[(a)]
    \item \label{e1a:sample}
    by taking a thousand samples from the stationary distribution, performing gradient flow on them, and seeing what percentage of samples ends up in each well;
    \item \label{e1a:zero} 
    by taking the zeroth order approximation $f_i = \exp(-\beta E_\theta(x_i))/\sum_j\exp(-\beta E_\theta(x_j))$ where $f_i$ is the approximation of the size of the the well at $x_i$;
    \item \label{e1a:2nd}
    by taking the second order approximation: 
\end{enumerate}
\begin{align}
    s_i &= \alpha \frac{\exp(-\beta E_\theta(x_i))}{\rho_i},\\
    \alpha &=  \Big(\sum_j \frac{\exp(-\beta E_\theta(x_j))}{\rho_j}\Bigr)^{-1},\\
    \rho_i(x_i) &= ((2\pi)^d \det (\beta^{-1} \dif^2 E_\theta(x_i)\inv))^{-1/2},
\end{align}
where $\dif^2 E_\theta(x_i)$ is the Hessian of the energy function at $x_i$, and $d$ is the dimension of the latent space. 

We report the $\ell^1$-distance to $(1/3, 1/3, 1/3)$ of each approximation for the top five best performing models ranked by second order approximation performance for both the underdamped and overdamped models in \Cref{tab:states:1}. For a more easily interpretable representation of the results, we also report the first coordinate of the learned distribution in \Cref{tab:coord:1}.

\begin{table*}[!h]
    \centering
    \begin{tabular}{l|c|c|c}
         & From sampling & $0^{\text{th}}$ order & $2^{\text{nd}}$ order\\
         \hline
         Underdamped & $0.23 \pm 0.040$ & $0.24\pm0.049$ & $0.22\pm0.0061$
    \end{tabular}
    \caption{$\ell^1$ distance between estimated distribution of states and actual distribution of states in experiment 2.}
    \label{tab:states:2}
\end{table*}

\begin{table*}[!h]
    \centering
    \begin{tabular}{l|c|c|c}
         & From sampling & $0^{\text{th}}$ order & $2^{\text{nd}}$ order\\
         \hline
         Coordinate 0 &$0.096\pm 0.014$&$ 0.090\pm 0.009$&$ 0.093\pm 0.003$ \\
         Coordinate 1 &$0.373 \pm 0.047$& $0.383 \pm 0.053$& $0.379 \pm 0.034$\\
         Coordinate 2 & $0.532\pm 0.041$&$0.527\pm 0.060$& $0.529 \pm 0.033$ \\
    \end{tabular}
    \caption{The distributions learned by the models from \Cref{tab:states:2}.\label{tab:coord:2}}
\end{table*}

\begin{table*}[!h]
    \centering
    \begin{tabular}{l|c|c}
         & mean accuracy & standard deviation\\
         \hline
         Underdamped & $85.3\% \pm 1.1\%$ & $4.2\% \pm 0.3\%$
    \end{tabular}
    \caption{Accuracy of unsupervised sequence segmentation for the same models as in \Cref{tab:states:2}\label{tab:seq-seg:2}}
\end{table*}

A second question is whether the discovered states in the trained prior dynamics coincide with the ground-truth latent states in the data-generating Markov chain, and whether the trained encoder network successfully maps unseen incoming data to the latent states investigated above; this is a measure of the degree to which the training has managed to model the whole dynamics. We generated 500 test sequences from the same three-state Markov chain and mapped them to the latent space using the encoder of  each of the models used for \Cref{tab:states:1}. Then we followed the gradient flow from each time step in each sequence of latent-space points to find out which well they belong to; this leads to a sequence of labels $(0,1,2)$. To compare these  labels to the ground truth, we selected the permutation of state labels $(0, 1, 2)$ for which the overlap is maximised. We then computed the mean and standard deviation of the accuracy over the 500 sequences. The results are shown in \Cref{tab:seq-seg:1}.

Note that the models operate in an online fashion resulting in the encoded latent state lagging slightly behind the true latent state. Nonetheless, these results clearly show that the wells in the energy landscape coincide with the discrete states behind the data. Moreover, the models were trained on sequences of length $200$, but are tested on sequences of length $500$ and clearly generalise well.

\subsection{Experiment 2}

\begin{figure}
    \centering
    \includemaybesvg[.5\textwidth]{../images/seqseg_online_a.pdf}{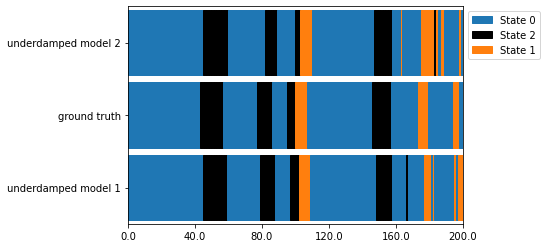}
    \caption{Results of unsupervised sequence segmentation by two models in experiment 2.\label{fig:seq-seg:2}}
\end{figure}
For the second experiment we restrict ourselves to underdamped models because those trained more stably in experiment 1. The distributions seem to represent the true distribution of the state much less accurately in this case, as shown in \Cref{tab:states:2} and \Cref{tab:coord:2}. This might be because these models were trained for much shorter due to time constraints. 

In this experiment we tested on 500 test sequences of length $200$. A visualisation of a segmented sequence in this experiment is given in \Cref{fig:seq-seg:2}. The permutations found here correspond to a sorting of the distribution from high to low, indeed matching the stationary distribution $(.45, .35, .2)$ behind the true states. So as can be seen in \Cref{tab:seq-seg:2}, the local minima do still correspond to the states underlying the data.

\section{Limitations and future research}
Langevin dynamics can not represent periodic behaviour. Filtering out periodic signals can be a feature. However, when this behaviour is not wanted, another term needs to be introduced in the drift. The addition of a divergence-free rotation field to the drift might alleviate this problem without sacrificing the interpretability of the energy landscape.

We had some trouble getting our NLD models to train consistently: the loss would become NaNs or jump several orders of magnitude. We conjecture that this is due to poor choice of hyper-parameters and architecture, since we encountered the same problems with the NSDE models we trained on the same data. Nonetheless, more work needs to be put into getting these models to train consistently.

\section{Conclusion}
To provide a more interpretable alternative to Neural Stochastic Differential Equations, we introduce Neural Langevin Dynamics, in which we replace the general drift term by the gradient of a trainable energy and let the system evolve by Langevin dynamics. The gain in interpretability comes both from the better options for visualisations of scalar functions over general vectors fields, and the better options for downstream processing in the form of classical methods from optimisation and analysis. With these methods, we can extract important properties such as stationary distributions and discrete states. When we apply NLD to example problems, it recovers the discrete states underlying the data. These results show promise that NLD is able to combine the performance of neural networks with the interpretability of classical differential equation based models.

\FloatBarrier
\bibliography{bibliography}
\end{document}